\documentclass[11pt,a4paper]{article}
\usepackage[utf8]{inputenc}
\usepackage[hyperref]{acl2020}
\usepackage{times}
\usepackage{latexsym}

\usepackage{microtype}
\usepackage{nameref}
\usepackage{latexsym}
\usepackage{natbib}
\usepackage{url}
\usepackage{booktabs}
\usepackage{tikz-qtree}
\usepackage{algpseudocode}
\usepackage{algorithm}
\usepackage{amsmath}
\usepackage{todonotes}
\usepackage{appendix}

\aclfinalcopy 
\def\aclpaperid{657} 

\usepackage{amsmath}

\title{Analyzing analytical methods:\\ The case of phonology in neural models of spoken language}

\author{Grzegorz Chrupała \\
  Cognitive Science and AI\\
  Tilburg University \\
  \texttt{g.chrupala@uvt.nl}\\\And
  Bertrand Higy\\ Cognitive Science and AI\\
  Tilburg University \\
  \texttt{b.j.r.higy@uvt.nl}\\\AND
  Afra Alishahi\\
  Cognitive Science and AI\\
  Tilburg University \\ \texttt{a.alishahi@uvt.nl}}

\date{}

\begin{document}
\maketitle
\begin{abstract}
Given the fast development of analysis techniques for NLP and speech
processing systems, few systematic studies have been conducted to
compare the strengths and weaknesses of each method.  As a step in
this direction we study the case of representations of phonology in
neural network models of spoken language. We use two commonly applied
analytical techniques, diagnostic classifiers and representational
similarity analysis, to quantify to what extent neural activation
patterns encode phonemes and phoneme sequences. We manipulate two
factors that can affect the outcome of analysis. First, we investigate
the role of learning by comparing neural activations extracted from
trained versus randomly-initialized models. Second, we examine the
temporal scope of the activations by probing both local activations
corresponding to a few milliseconds of the speech signal, and global
activations pooled over the whole utterance. We conclude that
reporting analysis results with randomly initialized models is
crucial, and that global-scope methods tend to yield more consistent
results and we recommend their use as a complement to local-scope
diagnostic methods.
\end{abstract}

\section{Introduction}
\label{sec:intro}
As end-to-end architectures based on neural networks became the tool of choice
for processing speech and language, there has been increased interest in
techniques for analyzing and interpreting the representations emerging in these
models. A large array of analytical techniques have been proposed and applied
to diverse tasks and architectures
\citep{belinkov2019analysis,alishahi2019analyzing}.

Given the fast development of analysis techniques for NLP and speech processing
systems, relatively few systematic studies have been conducted to compare the
strengths and weaknesses of each methodology and to assess the reliability and
explanatory power of their outcomes in controlled settings.  This paper reports
a step in this direction: as a case study, we examine the representation of
phonology in neural network models of spoken language. We choose three
different models that process speech signal as input, and analyze their learned
neural representations.

We use two commonly applied analytical techniques: (i) {\it diagnostic models}
and (ii) {\it representational similarity analysis} to quantify to what extent
neural activation patterns encode phonemes and phoneme sequences.

In our experiments, we manipulate two important factors that can affect the
outcome of analysis. One pitfall not always successfully avoided in work on
neural representation analysis is the {\it role of learning}. Previous work has
shown that sometimes non-trivial representations can be found in the activation
patterns of randomly initialized, untrained neural networks
\citep{DBLP:journals/corr/abs-1809-10040,chrupala-alishahi-2019-correlating}.
Here we investigate the representations of phonology in neural models of spoken
language in light of this fact, as extant studies have not properly controlled
for role of learning in these representations.

The second manipulated factor in our experiments is the {\it scope of the
extracted neural activations}. We control for the temporal scope, probing both
local activations corresponding to a few milliseconds of the speech signal, as
well as global activations pooled over the whole utterance.

When applied to global-scope representations, both analysis methods detect
a robust difference between the trained and randomly initialized
target models. However we find that in our setting, RSA applied to
local representations shows low correlations between phonemes and
neural activation patterns for both trained and randomly initialized
target models, and for one of the target models the local diagnostic
classifier only shows a minor difference in the decodability of
phonemes from randomly initialized versus trained network. This
highlights the importance of reporting analysis results with randomly
initialized models as a baseline.

This paper comes with a repository which contains instructions and
code to reproduce our experiments.\footnote{See \href{https://github.com/gchrupala/analyzing-analytical-methods}{https://github.com/gchrupala/analyzing-analytical-methods}.}

\section{Related work}
\label{sec:related}

\subsection{Analysis techniques}

Many current neural models of language learn representations that capture useful 
information about the form and meaning of the linguistic input.
Such neural representations are typically extracted from activations of various layers 
of a deep neural architecture trained for a target task such as automatic speech recognition 
or language modeling. 

A variety of analysis techniques have been proposed in the academic
literature to analyze and interpret representations learned by deep
learning models of language as well as explain their decisions; see
\citet{belinkov2019analysis} and \citet{alishahi2019analyzing} for a
review.
Some of the proposed techniques aim to  explain the behavior of a
network by tracking the response of individual or groups of neurons to
an incoming trigger \cite[e.g.,][]{nagamine2015exploring,krug2018neuron}.
In contrast, a larger body of work is dedicated to determining what type of linguistic
information is encoded in the learned representations. This type of analysis is the 
focus of our paper.
Two commonly used approaches to analyzing representations are:
\begin{itemize}
\item {\bf Probing techniques, or diagnostic classifiers,} i.e.\ methods which use the activations
from different layers of a deep learning architecture as input to a
prediction model \cite[e.g.,][]{adi2016fine,alishahi-etal-2017-encoding,hupkes2018visualisation,conneau-etal-2018-cram};
\item {\bf Representational Similarity Analysis (RSA)} borrowed from neuroscience
\cite{kriegeskorte2008representational} and used to correlate similarity
structures of two different representation spaces 
\cite{bouchacourt-baroni-2018-agents,chrupala-alishahi-2019-correlating,abnar-etal-2019-blackbox,abdou-etal-2019-higher}.
\end{itemize}
We use both techniques in our experiments to systematically compare their output.

\subsection{Analyzing random representations}

Research on the analysis of neural encodings of language has shown that in some cases, 
substantial information can be decoded from activation patterns of randomly initialized, untrained 
recurrent networks. It has been suggested that the dynamics of the network together with the characteristics
of the input signal can result in non-random activation patterns \citep{DBLP:journals/corr/abs-1809-10040}. 

Using activations generated by randomly initialized recurrent networks has a history in speech recognition and 
computer vision. Two better-known families of such techniques are called Echo State Networks (ESN) 
\citep{jaeger2001echo} and Liquid State Machines (LSM) \citep{maass2002real}. The general approach 
(also known as reservoir computing) is as follows: the input signal is passed through a randomly 
initialized network to generate a nonlinear response signal. This signal is then used as input to train a model 
to generate the desired output at a reduced cost.


We also focus on representations from randomly initialized neural
models but do so in order to show how training a model changes the
information encoded in the representations according to our chosen
analysis methods.

\subsection{Neural representations of phonology}

Since the majority of neural models of language work with text rather than speech, the bulk of work on 
representation analysis has been focused on (written) word and sentence representations. However, a number
of studies analyze neural representations of phonology learned by models that receive a speech signal as their 
input.

As an example of studies that track responses of neurons to controled
input, \citet{nagamine2015exploring} analyze local representations
acquired from a deep model of phoneme recognition and show that both
individual and groups of nodes in the trained network are selective to
various phonetic features, including manner of articulation, place of
articulation, and voicing.  \citet{krug2018neuron} use a similar
approach and suggest that phonemes are learned as an intermediate
representation for predicting graphemes, especially in very deep
layers.

Others predominantly use diagnostic classifiers for phoneme and
grapheme classification from neural representations of speech.  In one
of the their experiments \citet{alishahi-etal-2017-encoding} use
a linear classifier to predict phonemes from local activation patterns
of a grounded language learning model, where images and their spoken
descriptions are processed and mapped into a shared semantic
space. Their results show that the network encodes substantial
knowledge of phonology on all its layers, but most strongly on the
lower recurrent layers.

Similarly, \citet{belinkov2017analyzing} use diagnostic classifiers to
study the encoding of phonemes in an end-to-end ASR system with
convolutional and recurrent layers, by feeding local (frame-based)
representations to an MLP to predict a phoneme label.  They show that
phonological information is best represented in lowest input and
convolutional layers and to some extent in low-to-middle recurrent
layers.  \citet{belinkov2019analyzing} extend their previous work to
multiple languages (Arabic and English) and different datasets, and
show a consistent pattern across languages and datasets where both
phonemes and graphemes are encoded best in the middle recurrent
layers.

None of these studies report on phoneme classification from randomly
initialized versions of their target models, and none use global (i.e., utterance-level) 
representations in their analyses.

\section{Methods}
\label{sec:methods}
In this section we first describe the speech models which are the
targets of our analyses, followed by a discussion of the methods used
here to carry out these analyses.

\subsection{Target models}
We tested the analysis methods on three target models trained on speech
data.
\label{sec:target_models}
\paragraph{Transformer-ASR model}
\label{sec:trans-asr}
The first model is a transformer model
\citep{vaswani_attention_2017} trained on the automatic speech
recognition (ASR)
task. More precisely, we used a pretrained joint CTC-Attention
transformer model from the ESPNet toolkit \citep{watanabe2018espnet},
trained on the Librispeech dataset
\citep{panayotov_librispeech:_2015}.\footnote{We used ESPnet code from
  commit
  \href{https://github.com/espnet/espnet/commit/8fdd8e96b0896d97a63ab74ceb1cbc01c7652778}{8fdd8e9}
  with the pretrained model available from
  \href{https://tinyurl.com/r9n2ykc}{tinyurl.com/r9n2ykc}.}
The architecture is based on the hybrid CTC-Attention decoding scheme
presented by \citet{watanabe_hybrid_2017} but adapted to the transformer
model. The encoder is composed of two 2D convolutional layers
(with stride 2 in both time and frequency) and a linear layer,
followed by 12 transformer layers, while the decoder has 6 such
layers. The convolutional layers use 512 channels, which is also the
output dimension of the linear and transformer layers. The dimension
of the flattened output of
the two convolutional layers (along frequencies and channel) is then
20922 and 10240 respectively: we omit these two layers in our analyses
due to their excessive size.
The input to the model is made of a spectrogram with 80 coefficients
and 3 pitch features, augmented with the
SpecAugment method \citep{Park2019}. The output is
composed of 5000 SentencePiece subword tokens
\citep{kudo-richardson-2018-sentencepiece}. The model is trained for 120 epochs
using the optimization strategy from
\citet{vaswani_attention_2017}, also known as Noam optimization.
Decoding is performed with a beam of
size 60 for reported word error rates (WER) of 2.6\% and 5.7\% on the test
set (for the \texttt{clean} and \texttt{other} subsets respectively).

\paragraph{RNN-VGS model}
\label{sec:rnn-vgs}

The Visually Grounded Speech (VGS) model is trained on the task of
matching images with their corresponding spoken captions, first introduced
by \citet{harwath2015deep} and \citet{harwath2016unsupervised}. We
use the architecture of \citet{Merkx2019} which implemented
several improvements over the RNN model of \citet{chrupala-etal-2017-representations},
and train it on the Flickr8K Audio Caption Corpus
\citep{harwath2015deep}.  The speech encoder consists of one
1D convolutional layer (with 64 output channels) which
subsamples the input by a factor of two, and four bidirectional GRU
layers (each of size 2048) followed by a self-attention-based pooling
layer. The image encoder uses features from a pre-trained ResNet-152
model \citep{he2016deep}
followed by a linear projection. The loss function is a margin-based
ranking objective. Following \citet{Merkx2019} we trained the model
using the Adam optimizer \citep{kingma2014adam} with a cyclical learning
rate schedule \citep{smith2017cyclical}. The
input are MFCC features with total energy and delta and double-delta
coefficients with combined size 39.

\paragraph{RNN-ASR model}
\label{sec:rnn-asr}

This model is a middle ground between the two previous ones. It is
trained as a speech recognizer similarly to the transformer model but
the architecture of the encoder follows the RNN-VGS model
(except that the recurrent layers are one-directional in order to fit
the model in GPU memory).  The last GRU layer of the encoder is fed to
the attention-based decoder from \citet{bahdanau_neural_2015}, here
composed of a single layer of 1024 GRU units. The model is trained
with the Adadelta optimizer \citep{zeiler_adadelta:_2012}. The input
features are identical to the ones used for the VGS model; it is also
trained on the Flickr8k dataset spoken caption data, using the
original written captions as transcriptions.  The architecture of this model is not
optimized for the speech recognition task: rather it is designed to be
as similar as possible to the RNN-VGS model while still performing
reasonably on speech recognition (WER of 24.4\% on Flickr8k validation set with a beam of size 10).

\subsection{Analytical methods}
\label{sec:analytical}
We consider two analytical approaches:

\begin{itemize}
\item {\bf Diagnostic model} is a simple, often linear, classifier or
regressor trained to predict some information of interest given neural
activation patterns. To the extent that the model successfuly decodes
the information, we conclude that this information is present in
the neural representations.
\item {\bf Representational similarity analysis (RSA)} is a second-order
approach where similarities between pairs of some stimuli are measured in
two representation spaces: e.g. neural activation pattern space and a space
of symbolic linguistic representations such as sequences of phonemes or syntax
trees \citep[see][]{chrupala-alishahi-2019-correlating}. Then the
correlation between these pairwise similarity measurements quantifies
how much the two representations are aligned.
\end{itemize}
The diagnostic models have trainable parameters while the
RSA-based models do not, except when using a trainable pooling
operation.

We also consider two ways of viewing activation patterns in hidden layers as
representations:
\begin{itemize}
\item {\bf Local representations} at the level of a single frame
or time-step;
\item {\bf Global representations} at the level of the whole
utterance.
\end{itemize}
Combinations of these two facets give rise to the following concrete analysis models.

\paragraph{Local diagnostic classifier.} We use single frames of input
(MFCC or spectrogram) features, or activations at a single timestep as input
to a logistic diagnostic classifier which is trained to predict the
phoneme aligned to this frame or timestep.

\paragraph{Local RSA.}
We compute two sets of similarity scores. For neural representations,
these are cosine similarities
between neural activations from pairs of frames. For phonemic
representations our similarities are binary, indicating whether a pair
of frames are labeled with the same phoneme. Pearson's $r$
coefficient computed against  a binary variable, as in our setting, is
also known as point biserial correlation.

\paragraph{Global diagnostic classifier.}
We train a linear diagnostic classifier to predict the presence of phonemes
in an utterence based on global (pooled) neural activations.
For each phoneme $j$ the predicted probability that it is present in the
utterance with representation $\mathbf{h}$ is denoted as
$\mathrm{P}(j|\mathbf{h})$ and computed as:
\begin{equation}
  \label{eq:global_diagnostic}
   \mathrm{P}(j|\mathbf{h}) =  \mathrm{sigmoid}(\mathbf{W}\mathrm{\text{Pool}}(\mathbf{h})+\mathbf{a})_j
\end{equation}
where $\mathrm{Pool}$ is one of the pooling function in Section~\ref{sec:pooling}.

\paragraph{Global RSA.}
We compute pairwise similarity scores between global (pooled; see Section~\ref{sec:pooling})
representations and measure Pearson's $r$ with the pairwise string
similarities between phonemic transcriptions of
utterances. We define string similarity as:
\begin{equation}
  \mathrm{sim}(a, b) = 1 - \frac{\mathrm{Levenshtein}(a, b)}{\max(|a|, |b|)}
\end{equation}
where $|\cdot|$ denotes string length and $\mathrm{Levenshtein}$ is
the string edit distance.

\subsubsection{Pooling}
\label{sec:pooling}

The representations we evaluate are sequential: sequences of input
frames, or of neural activation states. In order to
pool them into a single global representation of the whole utterance
we test two approaches.
\paragraph{Mean pooling.} We simply take the mean for
each feature along the time dimension.
\paragraph{Attention-based pooling.}  Here we use a simple
self-attention operation with parameters trained to optimize the score
of interest, i.e.\ the RSA score or the error of the diagnostic classifier.
The attention-based pooling operator performs a weighted average
over the positions in the sequence, using scalar weights. The
pooled utterance representation $\mathrm{\text{Pool}}(\mathbf{h})$ is defined as:
\begin{equation}
  \label{eq:pooling1}
  \mathrm{\text{Pool}}(\mathbf{h}) = \sum_{t=1}^N \alpha_t \mathbf{h}_t,
\end{equation}
with the weights $\mathbf{\alpha}$ computed as:
\begin{equation}
  \label{eq:pooling2}
  \alpha_t = \frac{\exp(\mathbf{w}^T\mathbf{h}_t)}{\sum_{j=1}^N \exp(\mathbf{w}^T\mathbf{h}_j)},
\end{equation}
  where $\mathbf{w}$ are learnable parameters, and
  $\mathbf{h}_t$ is an input or activation vector at position $t$.\footnote{Note that the visually grounded speech models of
\citet{chrupala-etal-2017-representations,chrupala-2019-symbolic,Merkx2019} use similar mechanisms to aggregate
the activations of the final RNN layer; here we use it as part of the
analytical method to pool any sequential representation of interest. A
further point worth noting is that we use scalar weights $\alpha_t$ and
apply a linear model for learning them in order to keep the analytic
model simple and easy to train consistently.}

\subsection{Metrics}
\label{sec:metrics}
For RSA we use Pearson's $r$ to measure how closely
the activation similarity space corresponds to the phoneme or phoneme
string similarity space. For the diagnostic classifiers we use the
relative error reduction (RER) over the majority class baseline to
measure how well phoneme information can be decoded from the
activations.
\paragraph{Effect of learning}
In order to be able to assess and compare how sensitive the different
methods are to the effect of learning on the activation patterns, it
is important to compare the score on the trained model to that on the
randomly initialized model; we thus always display the two jointly.
We posit that a desirable property of an analytical method is that it is
sensitive to the learning effect, and that the scores on trained
versus randomly initialized models are clearly separated.

\paragraph{Coefficient of partial determination}
Correlation between similarity structures of two representational
spaces can, in principle, be partly due to the fact that both these
spaces are correlated to a third space. For example, were we to get a
high value for global RSA for one of the top layers of the RNN-VGS
model, we might suspect that this is due to the fact that string
similarities between phonemic transcriptions of captions are
correlated to visual similarities between their corresponding images,
rather than due to the layer encoding phoneme strings. In order to
control for this issue, we can carry out RSA between two spaces while
controling for the third, confounding, similarity space. We do this by
computing the {\it coefficient of partial determination} defined as the
relative reduction in error caused by including variable $X$ in a linear
regression model for $Y$:
\begin{equation}
  R^2_{\text{partial}}(Y,X|Z) = \frac{e_{Y\sim Z}-e_{Y\sim X+Z}}
    {e_{Y\sim Z}}
\end{equation}
where $e_{Y \sim X+Z}$ is the sum squared error of the model with all
variables, and $e_{Y \sim Z}$ is the sum squared error of the model
with $X$ removed.
Given the scenario above with the confounding space being visual
similarity, we identify $Y$ as the pairwise similarities in phoneme
string space, $X$ as the similarities in neural activation space, and
$Z$ as similarities in the visual space. The visual similarities are
computed via cosine similarity on the image feature vectors
corresponding to the stimulus utterances.

\subsection{Experimental setup}
\label{sec:experimental}
All analytical methods are implemented in Pytorch \citep{NEURIPS2019_9015}. The diagnostic
classifiers are trained using Adam with learning rate schedule which is
scaled by 0.1 after 10 epochs with no improvement in accuracy.  We
terminate training after 50 epochs with no improvement.
Global RSA with attention-based pooling is trained using Adam for 60
epochs with a fixed learning rate (0.001).  For all trainable models
we snapshot model parameters after every epoch and report the results
for the epoch with best validation score.
In all cases we sample half of the available data for training (if
applicable), holding out the other half for validation.

\paragraph{Sampling data for local RSA.}
When computing RSA scores it is common practice in neuroscience
research to use the whole upper triangular part of the matrices
containing pairwise similarity scores between stimuli, presumably
because the number of stimuli is typically small in that setting. In
our case the number of stimuli is very large, which makes using all
the pairwise similarities computationally taxing. More importantly,
when each stimulus is used for computing multiple similarity scores,
these scores are not independent, and score distribution changes
with the number of stimuli. We therefore use an alternative procedure
where each stimulus is sampled without replacement and used only in a
single similarity calculation.

\section{Results}
\label{sec:results}

\begin{figure*}
  \centering
  \includegraphics[scale=0.65]{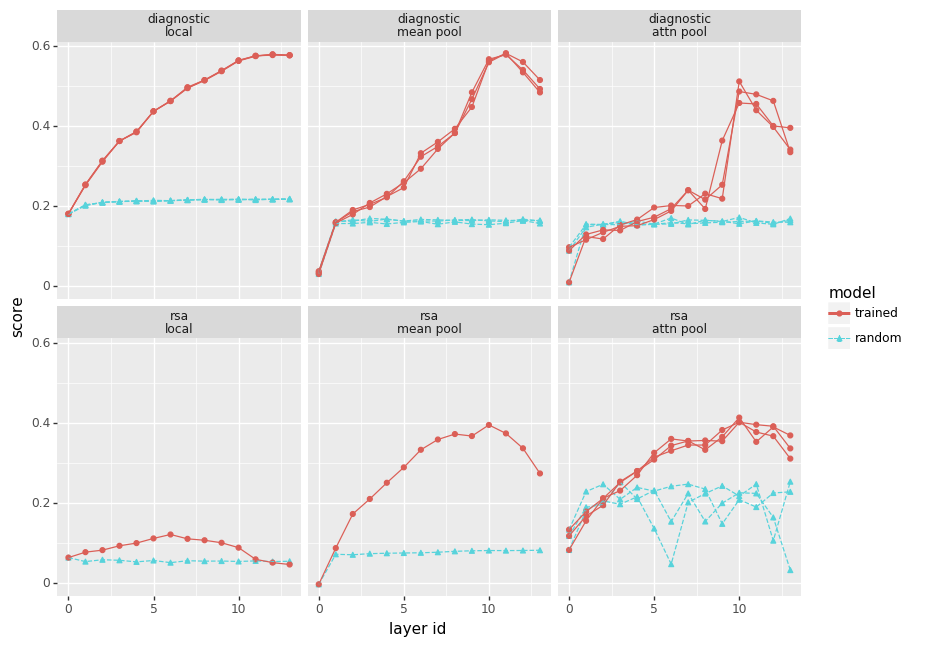}
  \caption{Results of diagnostic and RSA analytical methods applied to
    the Transformer-ASR model. The score is RER for the diagnostic methods and Pearson's $r$ for RSA.}
  \label{fig:results-trans-asr}
\end{figure*}

\begin{figure*}
  \centering
  \includegraphics[scale=0.65]{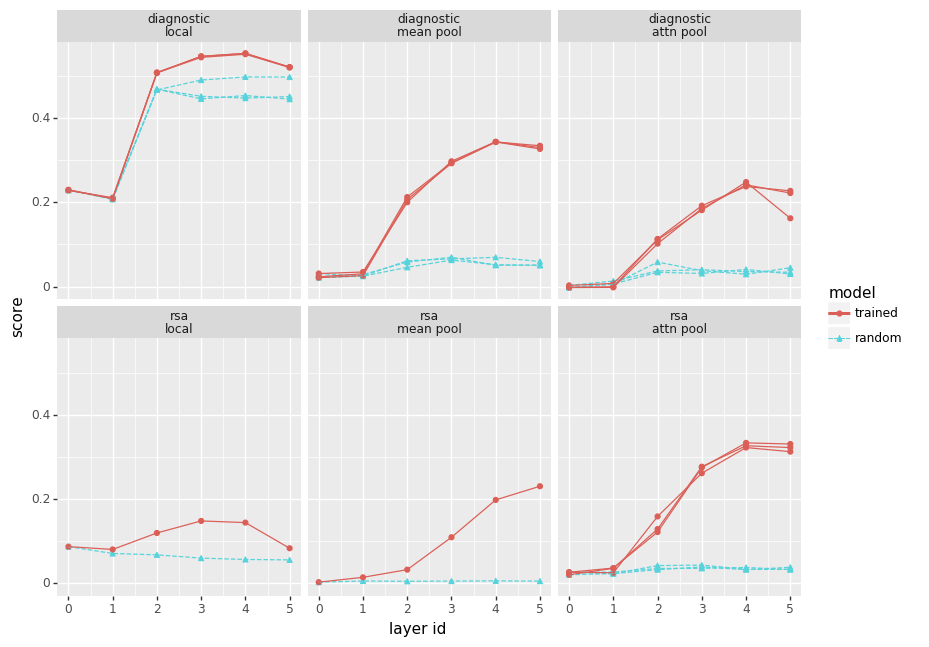}                                                                 
  \caption{Results of diagnostic and RSA analytical methods applied to
    the RNN-VGS model. The score is RER for the diagnostic methods and Pearson's $r$ for RSA. }
  \label{fig:results-rnn-vgs}
\end{figure*}

\begin{figure*}
  \centering
  \includegraphics[scale=0.65]{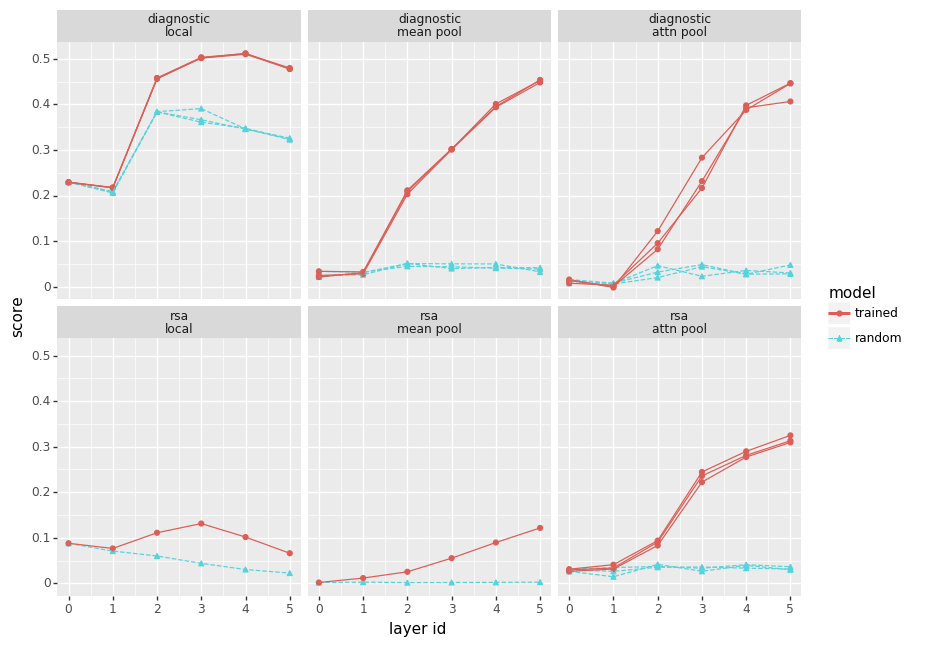}                                                                 
  \caption{Results of diagnostic and RSA analytical methods applied to
    the RNN-ASR model. The score is RER for the diagnostic methods and Pearson's $r$ for RSA. }
  \label{fig:results-rnn-asr}
\end{figure*}

Figures~\ref{fig:results-trans-asr}--\ref{fig:results-rnn-asr} display
the outcome of analyzing our target models. All three figures are
organized in a $2\times 3$ matrix of panels, with the top row showing
the diagnostic methods and the bottom row the RSA methods; the first
column corresponds to local scope; column two and three show global
scope with mean and attention pooling respectively.  The data points
are displayed in the order of the hierarchy of layers for each
architecture, starting with the input (layer id = 0).  In all the
reported experiments, the score of the diagnostic classifiers
corresponds to relative error reduction (RER), whereas for RSA we show
Pearson's correlation coefficient. For methods with trainable
parameters we show three separate runs with different random seeds in
order to illustrate the variability due to parameter initialization.

\begin{figure}[htb]
  \centering
  \includegraphics[scale=0.45]{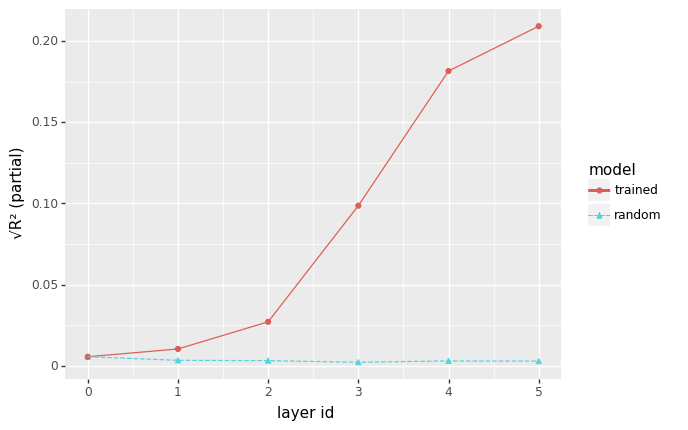}
  \caption{Results of global RSA with mean pooling on the RNN-VGS
    model, while controling for visual similarity. The score reported
    is the square root of the absolute value of the coefficient of
    partial determination $R^2_{\text{partial}}$.}
  \label{fig:rnn-vgs-r2_partial}
\end{figure}

Figure~\ref{fig:rnn-vgs-r2_partial} shows the results of global RSA
with mean pooling on the RNN-VGS target model, while
controling for visual similarity as a confound.

We will discuss the patterns of results observed for each model separately in the following sections.

\subsection{Analysis of the Transformer-ASR model}
\label{sec:results-trans-asr}

As can be seen in Figure~\ref{fig:results-trans-asr}, most reported experiments (with the exception of 
the local RSA) suggest that phonemes are best encoded in pre-final layers of the deep network. 
The results also show a strong impact of learning on the predictions of the analytical methods, 
as is evident by the difference between the performance using representations of the trained 
versus randomly initialized models. 

Local RSA shows low correlation values overall, and does not separate the trained versus random conditions well. 

\subsection{Analysis of the RNN-VGS model}
\label{sec:results-rnn-vgs}

Most experimental findings displayed in
Figure~\ref{fig:results-rnn-vgs} suggest that phonemes are best
encoded in RNN layers 3 and 4 of the VGS model.
They also show that the representations extracted from the trained model encode
phonemes more strongly than the ones from the random version of the
model.

However, the impact of learning is more salient with 
global than local scope: the scores of both local classifier and local 
RSA on random vs.\ trained representations are close to each other for all layers. For the 
global representations the performance on trained representations quickly diverges 
from the random representations from the first RNN layer onward.

Furthermore, as demonstrated in Figure~\ref{fig:rnn-vgs-r2_partial},
for top RNN layers of this architecture, the correlation between
similarities in the neural activation space and the similarities in
the phoneme string space is not solely due to both being correlated to
visual similarities: indeed similarities in activation space contribute
substantially to predicting string similarities, over and above the
visual similarities.

\subsection{Analysis of the RNN-ASR model}
The overall qualitative patterns for this target model are the same as
for RNN-VGS. The absolute scores for the global diagnostic variants
are higher, and the curves steeper, which may reflect that the
objective for this target model is more closely aligned with encoding
phonemes than in the case of RNN-VGS.

\subsection{RNN vs Transformer models}
In the case of the local diagnostic setting there is a marked contrast
between the behavior of the RNN models on the one hand and the
Transformer model on the other: the encoding of phoneme information
for the randomly initialized RNN is substantially stronger in the
higher layers, while for the randomly initialized Transformer the
curve is flat. This difference is likely due to the very different
connectivity in these two architectures.

With random weights in
RNN layer $i$, the activations at time $t$ are a function of the
features from layer $i-1$ at time $t$, mixed with the features from
layer $i$ at time $t-1$. There are thus effects of depth that may make
it easier for a linear diagnostic classifier to classify phonemes from
the activations of a randomly initialized RNN: (i) features are
re-combined among themselves, and (ii) local context features are also
mixed into the activations.

The Transformer architecture, on the other hand, does not have the
local recurrent connectivity: at each timestep $t$ the activations are
a combination of all the other timesteps and already in the first
layer, so with random weights, the activations are close to random,
and the amount of information does not increase with layer depth.

\begin{figure}[htb]
  \centering
  \includegraphics[scale=0.45]{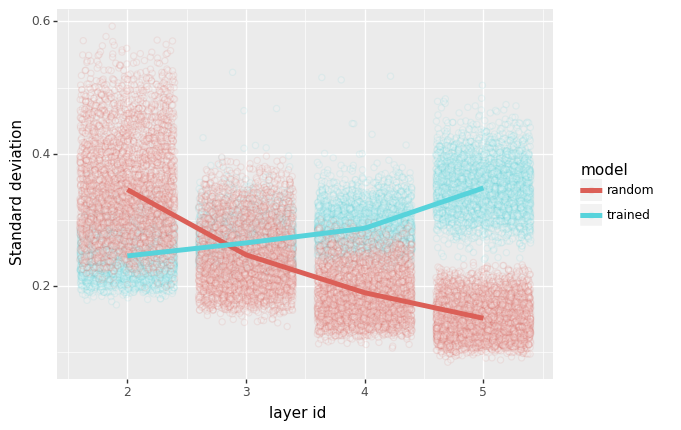}
  \caption{Standard deviation of pooled activations of the RNN layers for the RNN-VGS
    model.}
  \label{fig:std-rnn-vgs}
\end{figure}

In the global case, in the activations from random RNNs, pooling
across time has the effect of averaging out the vectors such that they
are around zero which makes them uninformative for the global
classifier: this does not happen to trained RNN
activations. Figure~\ref{fig:std-rnn-vgs} illustrates this point by
showing the standard deviations of vectors of mean-pooled activations
of each utterance processed by the RNN-VGS model for the randomly
initialized and trained conditions, for the recurrent
layers.\footnote{Only the RNN layers are show, as the different scale
  of activations in different layer types would otherwise obscure the
  pattern.}

\subsection{Summary of findings}
\label{sec:results-summary}

Here we discuss the impact of each factor in the outcome of our
analyses.
\paragraph{Choice of method.}
The choice of RSA versus diagnostic classifier interacts with scope,
and thus these are better considered as a combination. Specifically,
local RSA as implemented in this study shows only weak
correlations between neural activations and phoneme labels. It is
possibly related to the range restriction of point biserial
correlation with unbalanced binary variables. 

\paragraph{Impact of learning.}
Applied to the global representations, both analytical methods
are equally sensitive to learning. The results on random vs.\
trained representations for both methods start to diverge noticeably
from early recurrent layers. The separation for the local diagnostic
classifiers is weaker for the RNN models.

\paragraph{Representation scope.}
Although the temporal scale of the extracted representations
has not received much attention
and scrutiny, our experimental findings suggest that it is an
important choice. Specifically, global representations are more
sensitive to learning, and more consistent across different analysis
methods. Results with attention-based learned pooling are in general more
erratic than with mean pooling. This reflects the fact that analytical
models which incorporate learned pooling are more difficult to optimize
and require more careful tuning compared to mean pooling.

\subsection{Recommendations}
\label{sec:recommendations}
Given the above findings, we now offer tentative recommendations on
how to carry out representational analyses of neural models.
\begin{itemize}
\item Analyses on randomly initialized target models should be run as a baseline.
  Most scores on these models were substantially
  above zero: some relatively close to scores on trained models.
\item It is unwise to rely on a single analytical approach, even a
  widely used one such as the local diagnostic classifier. With solely this method
  we would have concluded that, in RNN models, learning has only a weak effect on the
  encoding of phonology.
\item Global methods applied to pooled representations should be
  considered as a complement to standard local diagnostic methods. In
  our experiments they show more consistent results.
\end{itemize}

\section{Conclusion}
\label{sec:conclusion}
In this systematic study of analysis methods for neural models of
spoken language we offered some suggestions on best practices in this
endeavor. Nevertheless our work is only a first step, and several
limitations remain. The main challenge is that it is often difficult
to completely control for the many factors of variation in the target
models, due to the fact that a particular objective function, or even
a dataset, may require relatively important architectural
modifications. In future we will sample target models with a larger
number of plausible combinations of factors. Likewise, a choice of an
analytical method may often entail changes in other aspects of the
analysis: for example, unlike a global diagnostic classifier, global
RSA captures the sequential order of phonemes.  In future work we hope
to further disentangle these differences.

\section*{Acknowledgements}
Bertrand Higy was supported by a NWO/E-Science Center grant
number~027.018.G03.

\bibliography{biblio,anthology}
\bibliographystyle{acl_natbib}

\end{document}